%% file: kim2017vis.tex
\documentclass{article}

\PassOptionsToPackage{numbers, compress}{natbib}
\usepackage[final]{nips_2017}

\usepackage[utf8]{inputenc} 
\usepackage[T1]{fontenc}    
\usepackage{hyperref}       
\usepackage{url}            
\usepackage{booktabs}       
\usepackage{amsfonts}       
\usepackage{nicefrac}       
\usepackage{microtype}      

\usepackage[pdftex]{graphicx}

\usepackage{amsmath}
\usepackage{stmaryrd}
\usepackage{dsfont}
\usepackage{algorithm}
\usepackage{algorithmic}
\usepackage{wrapfig}    

\include{mysymbols}

\title{Visual Explanations from Hadamard Product in Multimodal Deep Networks}

\author{
  Jin-Hwa Kim \\
  Seoul National University\\
  Seoul, Republic of Korea \\
  \texttt{jhkim@bi.snu.ac.kr} \\
  \And
  Byoung-Tak Zhang \\
  Seoul National University\\
  Seoul, Republic of Korea \\
  \texttt{btzhang@bi.snu.ac.kr} \\
}

\begin{document}

\maketitle

\begin{abstract}
The visual explanation of learned representation of models helps to understand the fundamentals of learning.
The attentional models of previous works used to visualize the attended regions over an image or text using their learned weights to confirm their intended mechanism.
Kim et al. (2016) show that the Hadamard product in multimodal deep networks, which is well-known for the joint function of visual question answering tasks, implicitly performs an attentional mechanism for visual inputs.
In this work, we extend their work to show that the Hadamard product in multimodal deep networks performs not only for visual inputs but also for textual inputs simultaneously using the proposed gradient-based visualization technique.
The attentional effect of Hadamard product is visualized for both visual and textual inputs by analyzing the two inputs and an output of the Hadamard product with the proposed method and compared with learned attentional weights of a visual question answering model.
\end{abstract}

\section{Introduction}

As a multimodal joint function, Hadamard product is widely used in the multimodal learning tasks. Many state-of-the-art models used the Hadamard product as a joint function to achieve competitive performance~\cite{Kim2016b,Kim2017,Nam2016,Teney2017} for the visual question answering (VQA)~\cite{agrawal2017vqa}, and one of them won the recent VQA challenge~\cite{Teney2017}. 

The characteristic of Hadamard product is studied in deep neural networks. MI-RNN~\cite{Wu2016a} uses this to integrate different information flows within RNN. They show that hidden activations are not saturated toward $\pm1.0$ comparing to the addition, which implies that the gradient of \textit{tanh} function is not vanished. For the MLB~\cite{Kim2017}, they show that Hadamard product performs low-rank bilinear pooling in deep neural networks.

In this paper, we show that the analysis of the input and output of Hadamard product in multimodal deep networks is sufficient to visualize the cross-grounding between two modalities. Unlike the previous works, this method does not need an annotated label nor attentional weights. These results suggest that Hadamard product as a multimodal joint function gives not only excellent performance in the task but also visual explanations for the vision-language, input modalities.

\section{Previous Works}

Class Activation Mapping (CAM) is proposed to identify discriminative regions in image classification tasks~\cite{Zhou2016}. CAM utilizes global average pooling layers to get the representations to localize the regions. However, it has the limitation of the CNN architecture and the modification of architecture requires re-training. Grad-CAM~\cite{Selvaraju2016} generalizes this to various CNN models. The Grad-CAM can also visualize in the other tasks, \eg~image captioning and visual question answering, with the similar approach. Unlike the previous methods, our method is an unsupervised visualization, and a direct visualization of vision-language cross-groundings occurred in the joint function.

\section{Visual Explanations from Hadamard Product}

In their work~\cite{Kim2016b}, they visualize the difference between the intermediate visual input $\mathcal{V}_i$ and the output of Hadamard product $\mathcal{F}_i$ between the intermediate visual input $\mathcal{V}_i$ and the intermediate textual input $\mathcal{Q}_i$, in the space of image for three layers using standard back-propagation. \begin{gather}
  \frac{\partial{\mathcal{L}_{i}}}{\partial{\mathcal{I}}} = 
    \frac{\partial{  \big( \frac{1}{2}||\mathcal{V}_i-\mathcal{F}_i||_2^2 \big)  }}{\partial{\mathcal{I}}} = 
    \frac{\partial{\mathcal{V}_i}}{\partial{\mathcal{I}}}(\mathcal{V}_i-\mathcal{F}_i)
\end{gather}
where $\mathcal{I}$ is an input image to ResNet-152, and $\mathcal{V}_i$ is a function of $\mathcal{I}$; however, though $\mathcal{F}_i$ is also a function of $\mathcal{V}_i$, the $\mathcal{F}_i$ is treated as a constant for the visualization purpose. 
We speculate that the constant $\mathcal{F}_i$ set the virtual target of joint representation and the mean squared error between $\mathcal{F}_i$ and $\mathcal{V}_i$ indicates the amount of deviation of $\mathcal{V}_i$ from $\mathcal{F}_i$. 
Empirically, this way of visualization is more effective than not fixing the $\mathcal{F}_i$.

We define the visual explanation of vision and language using the inputs to Hadamard product and the output of the product:
\begin{align}
  \nabla_v := (\mathcal{V}-\mathcal{F}) \partial{\mathcal{V}} / \partial{\mathcal{I}} \label{eq:vis_v} \\
  \nabla_q := (\mathcal{Q}-\mathcal{F}) \partial{\mathcal{Q}} / \partial{q} \label{eq:vis_q}
\end{align}
Unlike the their work, we use \textit{guided back-propagation} for ReLU activations, which uses the only positive gradient of output for back-propagation, for a better visualization. This imputed version of gradient is calculated using the ResNet-152, the feature extractor. Equation~\ref{eq:vis_q} is newly introduced with the same gist. Here, the $q$ is the embedded word vectors for a given question which are looked up by the indices of tokens. Notice that the dimensions of $\mathcal{V}$, $\mathcal{Q}$, and $\mathcal{F}$ are the same since $\mathcal{F}$ is the output of element-wise multiplication of $\mathcal{V}$ and $\mathcal{Q}$. Moreover, these definitions can be generalized to the other models which uses Hadamard product as multimodal joint function. 

\section{Experiments}

\setlength{\columnsep}{1.5em}
\begin{wrapfigure}{r}{0.33\textwidth}
  \centering
  \includegraphics[width=\linewidth]{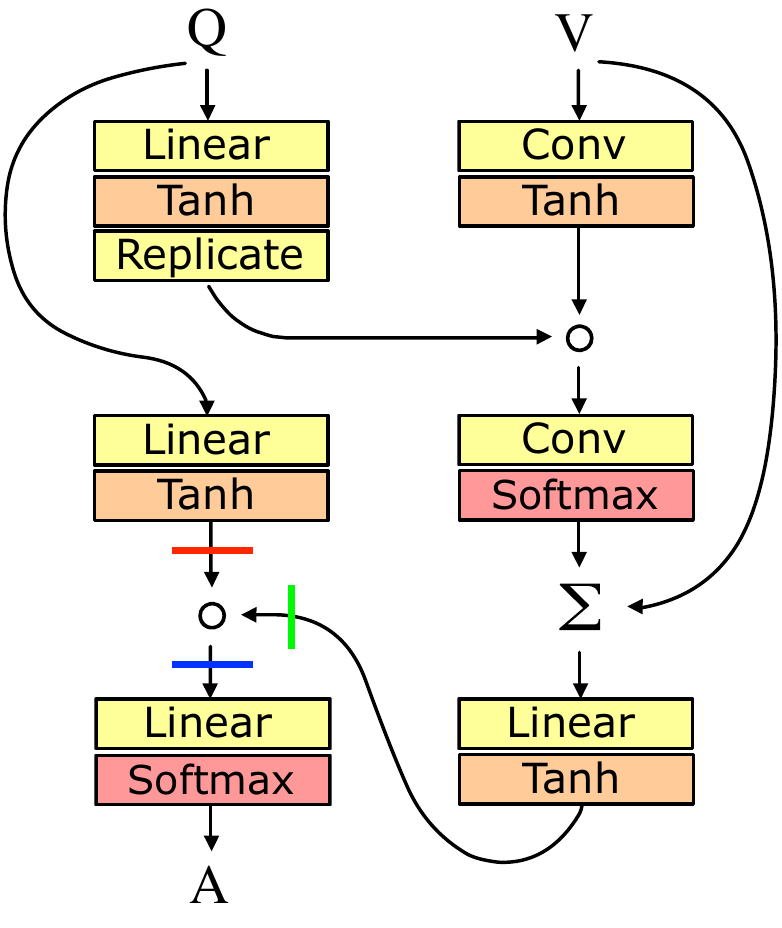}
  \caption{A diagram of MLB indicating the analyzing points; \textbf{red}: textual input $\mathcal{Q}$, \textbf{green}: visual input $\mathcal{V}$, and \textbf{blue}: the output of Hadamard product $\mathcal{F}$. \label{fig:schema}}
  \vspace{-2em}
\end{wrapfigure}

We use the multimodal low-rank bilinear attention networks (MLB)~\cite{Kim2017} as the VQA model for visualization, to compare the visual explanation for visual input with attentional weights, and to show the visual explanation for textual input. The MLB provides an efficient attention mechanism for visual question-answering tasks, based on the interpretation of Hadamard product as a key operator for low-rank bilinear pooling.  

\subsection{Multimodal Low-rank Bilinear Attention Networks}
\label{sec:mlb}

The inputs are a question embedding vector $\vq$, which is the output of a learnable Skip-thought Vectors model~\cite{Kiros2015}, and a set of visual feature vectors $\mF$ over $S \times S$ lattice space, which is the output of the fixed ResNet-152 model~\cite{He2015}. In this section, we briefly describe the structure of MLB.

Attention mechanism uses an attention probability distribution $\alpha$ over $S \times S$ lattice space. Here, using low-rank bilinear pooling, $\alpha$ is defined as:  \begin{align}
   \alpha &= \softmax \Big(\mP_\alpha^T \big( \sigma(\mU^T_{\vq} \vq \cdot \mathds{1}^T) \circ \sigma(\mV^T_{\mF} \mF^T) \big) \Big) \label{eq:alpha}
\end{align}
where $\alpha \in \R^{G \times S^2}$, $\mP_\alpha \in \R^{d \times G}$, $\sigma$ is a hyperbolic tangent function, $\mU_\vq \in \R^{N \times d}$, $\vq \in \R^N$, $\mathds{1} \in \R^{S^2}$, $\mV_\mF \in \R^{M \times d}$, and $\mF \in \R^{S^2 \times M}$. If $G > 1$, multiple glimpses are explicitly expressed as in \citet{Fukui2016}, conceptually similar to \citet{Jaderberg2015}. And, the $\softmax$ function applies to each row vector of $\alpha$. The bias terms are omitted for simplicity. 

Attended visual feature $\hat{\vv}$ is a linear combination of $\mF_i$ with the corresponding coefficients $\alpha_{g,i}$. Each attention probability distribution $\alpha_{g}$ is for a glimpse $g$. For $G > 1$, $\hat{\vv}$ is the concatenation of resulting vectors $\hat{\vv}_g$ as: \begin{align} 
  \hat{\vv} &= \bigparallel_{g=1}^G \sum_{s=1}^{S^2} \alpha_{g,s} \mF_{s} \label{eq:concat}
\end{align}
where $\bigparallel$ denotes concatenation of vectors. The posterior probability distribution is an output of a $\softmax$ function, whose input is the result of another low-rank bilinear pooling of $\vq$ and $\hat{\vv}$ as: \begin{gather}
   \mathcal{Q} := \sigma(\mW_\vq^T\vq), \hspace{1em}
   \mathcal{V} := \sigma(\mV_{\hat{\vv}}^T\hat{\vv}), \hspace{1em} 
   \mathcal{F} := \mathcal{Q} \circ \mathcal{V} \\
   p(a|\vq,\mF;\Theta) = \softmax \big( \mathds{\mP}_o^T \mathcal{F} \big) \\
  \hat{a} = \argmax_{a \in \Omega} p(a|\vq,\mF;\Theta)
\end{gather}
where $\hat{a}$ denotes a predicted answer, $\Omega$ is a set of candidate answers and $\Theta$ is an aggregation of entire model parameters.
Our method uses the intermediate representations, $\mathcal{Q}$, $\mathcal{V}$, and $\mathcal{F}$.

In Figure~\ref{fig:schema} indicates the analyzing points, $\mathcal{Q}$, $\mathcal{V}$, and $\mathcal{F}$. The \textit{Replicate} module copies an question embedding vector to match with $S^2$ visual feature vectors. \textit{Conv} modules indicate $1 \times 1$ convolution to project channel dimension, which is computationally equivalent to linear projection for the channel.

\subsection{Post-processing}

Using Equation~\ref{eq:vis_v} and \ref{eq:vis_q}, we get the gradients of $\nabla_v \in \R^{C \times H \times W}$ and $\nabla_q \in \R^{\rho \times D}$. The $\nabla_v$ has the same size of a RGB image and the $\rho$ is the number of tokens in a question, and $D$ is the dimension of word embedding vector. Then, each pixel $\nabla_{v(i,j)}$ is normalized (channel-wise) by:\begin{align} 
  \hat{\nabla}_{v(i,j)} &= (\nabla_{v(i,j)} - \mu(\nabla_{v})) / \sigma(\nabla_{v})
\end{align}
where $\mu$ denotes mean and $\sigma$ denotes standard deviation. 

For the question, we take the absolute values of $\nabla_{q}$, followed by the summation over $D$ for the $i$-th token:
\begin{align}
  \hat{\nabla}_{q(i)} = \sum_{d=1}^D |\nabla_{q(i,d)}|
\end{align}
Then, the standard score $z_i$ is calculated to get relative importance of each token as follows: \begin{align}
  z_i = \big( \hat{\nabla}_{q(i)} - \mu(\hat{\nabla}_{q}) \big) / \sigma(\hat{\nabla}_{q}).
\end{align}

\begin{figure}[ht]
  \centering
  \includegraphics[width=0.7\linewidth]{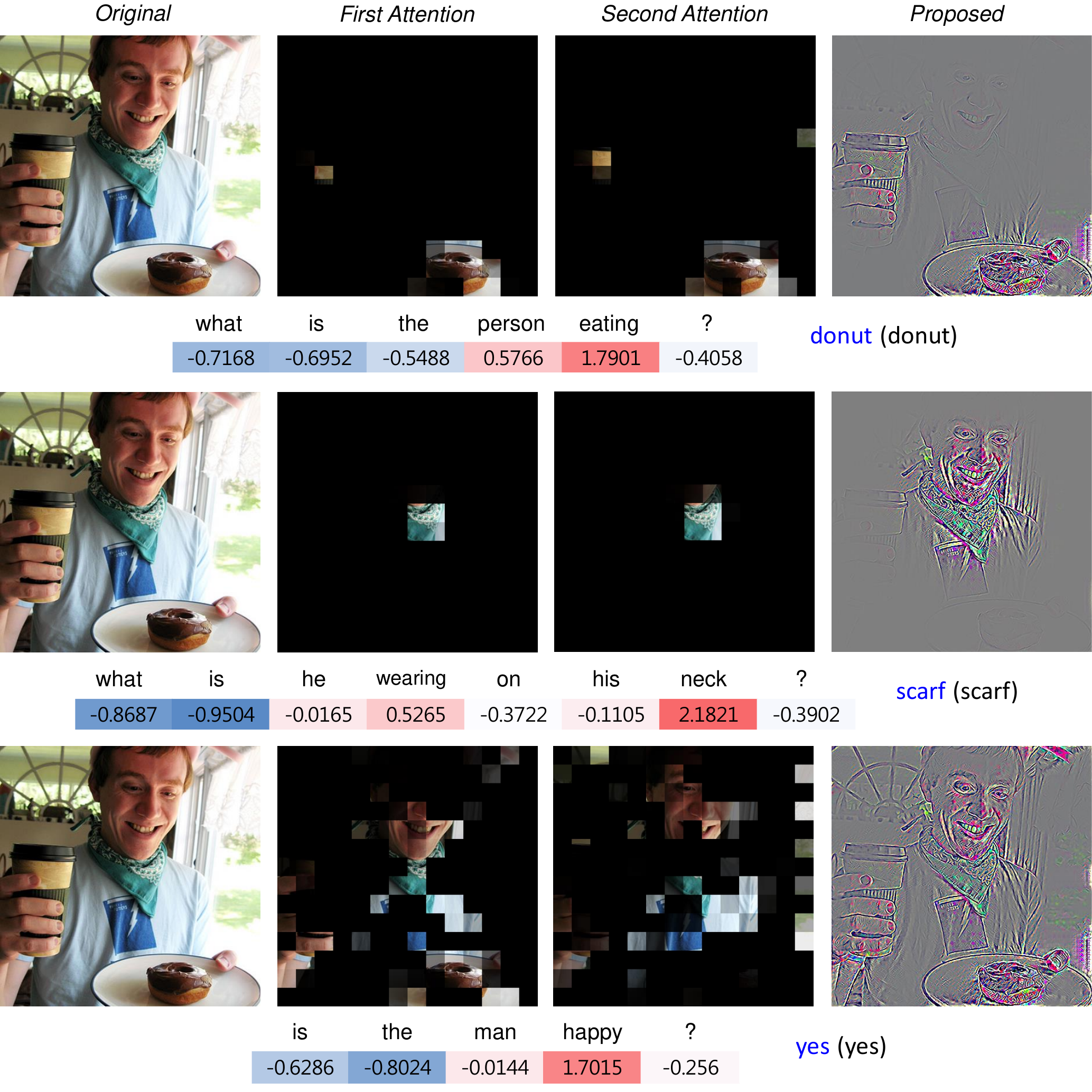}
  \caption{The visualization of attentional weights $\alpha_g$ (second and third columns), visual explanations for visual (forth column), and textual (bottom of each row) inputs. The two attention maps are represented on the $14 \times 14$ lattice space, while the proposed method represents an attended region on the image pixels. The plate of donut is visualized in the proposed method (first row), and the scarf (second row).}
  \label{fig:vis}
\end{figure}

\section{Results and Discussions}

Figure~\ref{fig:vis} shows an example of the visual explanations. The first column shows an input image, and the second and third columns show the first and second attention maps representing $\alpha_1$ and $\alpha_2$ (MLB uses the G of 2). Notice that $\hat{\vv}$ represents the concatenation of $g$ attended visual features. Through this, the model learns to generate appropriate attention probability distributions $\alpha_g$ in parallel. The first and second rows show that $\alpha_1$ and $\alpha_2$ has similar distributions; however, the third row shows some difference.

The fourth column shows the visualization of our proposed method. Although this visualization comes from the analysis of Hadamard product, it shows a similar result to the attention maps. The first and second attend to \textit{a donut on a plate} and \textit{a scarf in the neck}, respectively. In the third row, the proposed visualization seems to represent both of the first and second attention maps in an additive way.

The visual explanation of textual input shows a plausible result that the nouns are significantly attended (red), whereas \textit{wh} words, verbs, adjectives, and articles are less attended (blue) in the bottom of each row. Take-home message is two-fold: \textbf{1)} these results are competitive with the explicit textual attention models~\cite{Nam2016,Lu2016}. The explicit textual attention of theirs are not giving unprecedented attention to the text (notice that Hadamard product is a well-known joint function in the VQA tasks), rather it might be working as the regularization using selective weights. \textbf{2)} there is no sufficient evidence that the visual attention is based on the comprehension of a question. Because the textual attention is not expanding to the verbs and propositions connected to the nouns (\eg \textit{`on'} of \textit{`wearing on his neck'}), which phenomena seem to be consistent with the co-attention models~\cite{Nam2016,Lu2016}, although the work~\cite{Lu2016} tried to mitigate the problem using word, phrase, and question-level features.

\section{Conclusions}

In this work, we show that the Hadamard product in multimodal deep networks implicitly performs an attentional mechanism not only for visual inputs but also for textual inputs simultaneously using the proposed gradient-based visualization technique in a visual question answering model. 
Though this technique is based on the analysis of Hadamard product in multimodal deep networks, it shows competitive results with the explicit visualization of learned attentional weights.
Our results suggest that the explicit textual attention is not providing a unique attentional mechanism to the textual input. Instead, it might be the regularization using selective weights which are learned by training.
Moreover, we cautiously argue that textual attention is biased toward the noun words appeared in a given text which limits the inferential capability of the model.

{\small
\vspace{1em}\noindent\textbf{Acknowledgments.}
JK is supported by 2017 Google PhD Fellowship, and partly by NAVER Corp. and the Korea government (IITP-2017-0-01772-VTT, IITP-R0126-16-1072-SW.StarLab, KEIT-10060086-RISF).
}

\newpage

\bibliographystyle{plainnat}
\bibliography{kim2017hada,kim2017vis}

\clearpage

\setcounter{section}{0}
\renewcommand\theHsection{\Alph{section}}
\renewcommand\thesection{\Alph{section}}
\renewcommand\thesubsection{\thesection.\arabic{subsection}}

\section{Appendix}

\begin{figure}[h]
  \centering
  \includegraphics[width=0.8\linewidth]{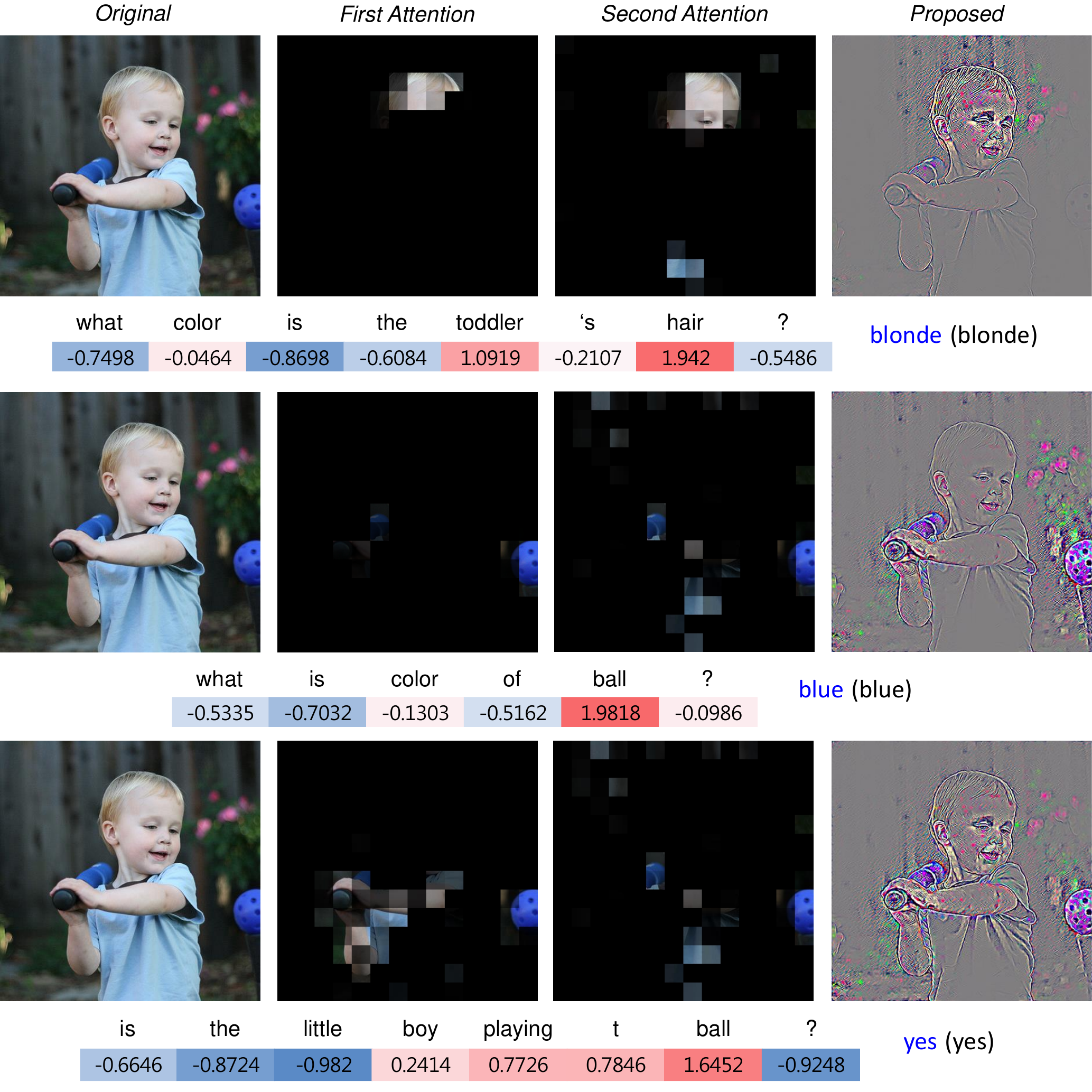}
  \caption{Another examples of the visualization.}
  \label{fig:more}
\end{figure}

\begin{figure}[t!]
  \centering
  \includegraphics[width=0.8\linewidth]{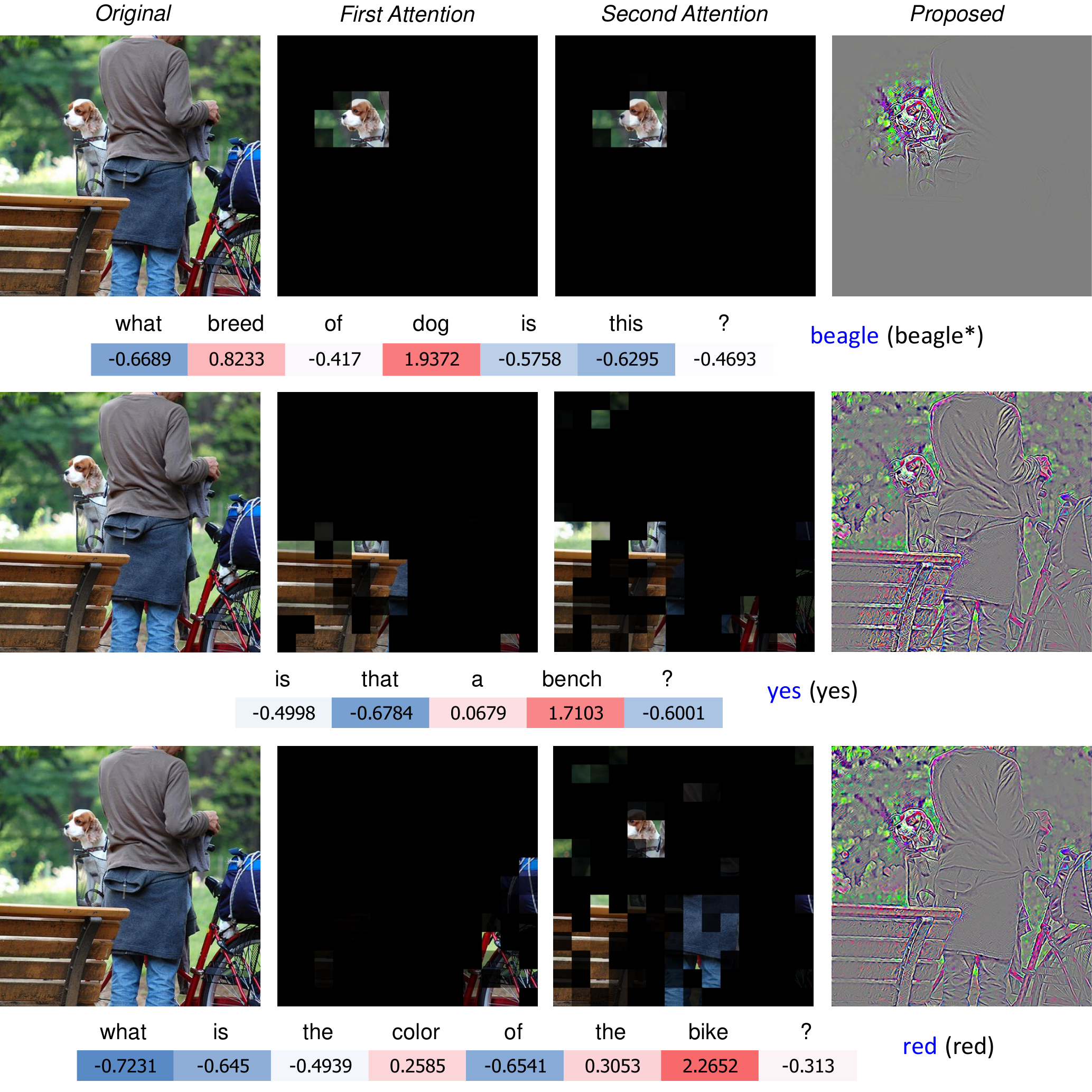}
  \caption{Another examples of the visualization. In the second and third rows of the forth column, the attended area of bench and bike is slightly different. Interestingly, the attention maps of the third row are different from each other, the first attention shows the part of bike, whereas the second attention shows the other salient objects. \textbf{*}Which is unclear to be Beagle or Charles Spaniel.}
  \label{fig:more1}
\end{figure}

\begin{figure}[ht]
  \centering
  \includegraphics[width=0.8\linewidth]{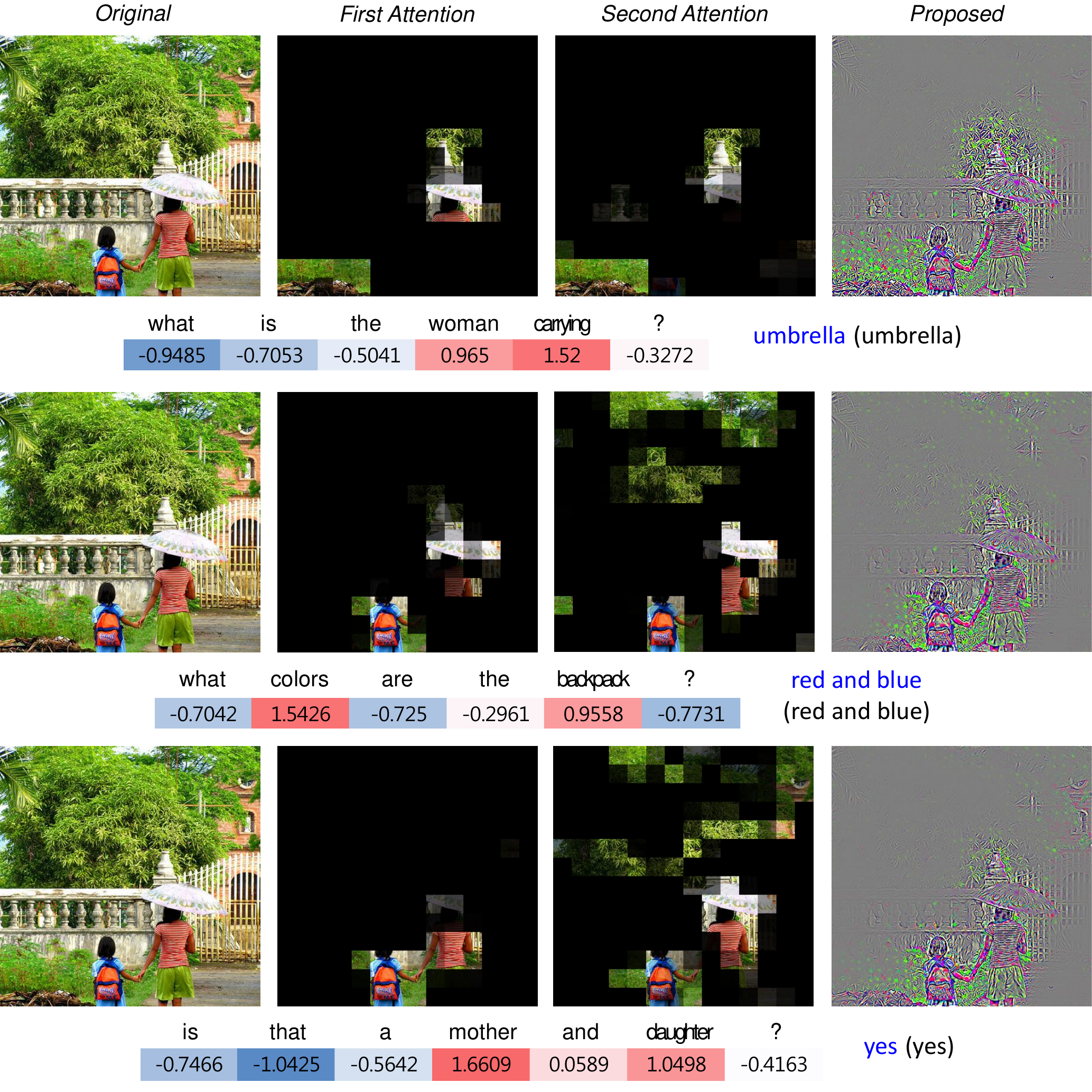}
  \caption{Another examples of the visualization. In the first and second rows of the fourth column, there is a very subtle difference between umbrella and backpack (distinguishable with two standard deviation threshold). The backpack has blue color in its side area (second row). We do not know the relationship between the two but only can infer from the pose of hand in hand.}
  \label{fig:more2}
\end{figure}

The supplementary examples of our method are shown in Figure~\ref{fig:more}, \ref{fig:more1}, and \ref{fig:more2}. This example emphasizes the importance of visual explanation. In the first row, the question is `\textit{what color is the toddler's hair?}' and the corresponding answer is `\textit{blonde}'. Without the visualization, we do not know whether the model is \textit{purely} biased from the data distribution or not. Although there is the possibility that the model is biased toward `\textit{blonde hair}' \textit{and} attends the hair in the given image, this visual explanation helps to assess the model.

\end{document}

%% file: mysymbols.tex
\newcommand{\vv}[0]{\mathbf{v}}

\newcommand{\vq}[0]{\mathbf{q}}

\newcommand{\mU}[0]{\mathbf{U}}
\newcommand{\mV}[0]{\mathbf{V}}
\newcommand{\mW}[0]{\mathbf{W}}
\newcommand{\mP}[0]{\mathbf{P}}
\newcommand{\mF}[0]{\mathbf{F}}

\newcommand{\R}[0]{\mathds{R}}

\usepackage{xspace}
\newcommand{\eg}[0]{\textit{e.g.}}

\DeclareMathOperator*{\argmax}{arg\,max}

\newcommand{\softmax}[0]{\text{softmax}}